# Estimating Lexical Priors for Low-Frequency Syncretic Forms

Harald Baayen
Max Planck Institute for Psycholinguistics
Wundtlaan 1, 6525 XD, Nijmegen, The Netherlands
`baayen@mpi.nl`

Richard Sproat
AT&T Bell Laboratories
600 Mountain Avenue
Murray Hill, NJ 07974, USA
`rws@research.att.com`

**Abstract**

Given a previously unseen form that is morphologically n-ways ambiguous, what is the best estimator for the lexical prior probabilities for the various functions of the form? We argue that the best estimator is provided by computing the relative frequencies of the various functions among the hapax legomena — the forms that occur exactly once in a corpus. This result has important implications for the development of stochastic morphological taggers, especially when some initial hand-tagging of a corpus is required: For predicting lexical priors for very low-frequency morphologically ambiguous types (most of which would not occur in any given corpus) one should concentrate on tagging a good representative sample of the hapax legomena, rather than extensively tagging words of all frequency ranges.

## 1 Introduction

In many inflected languages it is common to find two or more slots in an inflectional paradigm that are filled with the same form. This phenomenon, termed *syncretism*, can be illustrated by a Dutch example such as *lopen* 'walk', which can either be the infinitive form ('to walk'), or the plural present tense form ('we, you or they walk'). In some cases syncretism is completely systematic; for example in the case cited in Dutch, where the *-en* suffix can always function in the two ways cited, or in Latin, where the plural dative and ablative forms of nouns and adjectives are always identical, no matter what paradigm the noun belongs to. In other cases, a particular instance of syncretism may be displayed only in some paradigms; for example, Russian feminine nouns, such as *loshad'* 'horse' (Cyrillic лошадь), have the same form for both the genitive singular — *loshadi* (Cyrillic лошади) — and the nominative plural, whereas masculine nouns typically distinguish these forms.

Syncretism and related morphological ambiguities present a problem for statistical models of tagging since such models normally presume some estimate of the lexical prior for a given form (Church, 1988; DeRose, 1988; Kupiec, 1992). Assuming one has a tagged corpus, one can usually get reasonable estimates of the lexical priors for the frequent forms (such as *lopen* 'walk'), but for infrequent or unseen forms, it is less clear how to compute the estimate. So, consider another Dutch example like *aanlokken* 'entice, appeal'. This form occurs only once, as an infinitive, in the Uit den Boogaart (henceforth UdB) corpus (Uit den Boogaart, 1975); in other words it is a *hapax legomenon* in this corpus. Obviously the lexical prior probability of this form expressing the finite plural is not zero, the Maximum Likelihood Estimate (MLE) being a poor estimate in such cases. And when one considers forms that do not occur in the training corpus (e.g. *bedraden* 'to wire') the situation is even worse. The problem then is to provide a more reasonable estimate of the relative probabilities of the various potential functions of such forms.

## 2  Estimating the Lexical Priors for Rare Forms

For a common form like *lopen* 'walk' a reasonable estimate of the lexical prior probabilities is the MLE, computed over all occurrences of this form. So, in the UdB corpus, *lopen* occurs 92 times as an infinitive and 43 times as a finite plural, so the MLE estimate of the probability of the infinitive is 0.68. For low frequency forms like *aanlokken* or *bedraden*, one might consider basing the MLE on the aggregate counts of all ambiguous forms in the corpus. In the UdB corpus, there are 21703 infinitive tokens, and 9922 finite plural tokens, so the MLE for *aanlokken* being an infinitive would be 0.69. Note, however, that the application of this overall MLE presupposes that the relative frequencies of the various functions of a particular form is independent of the frequency of the form itself. For the Dutch example at hand, this presupposition predicts that if we were to classify *-en* forms according to their frequency, and then for each frequency class thus defined, plot the relative frequency of infinitives and finite plurals, the regression line should have a slope of approximately zero.

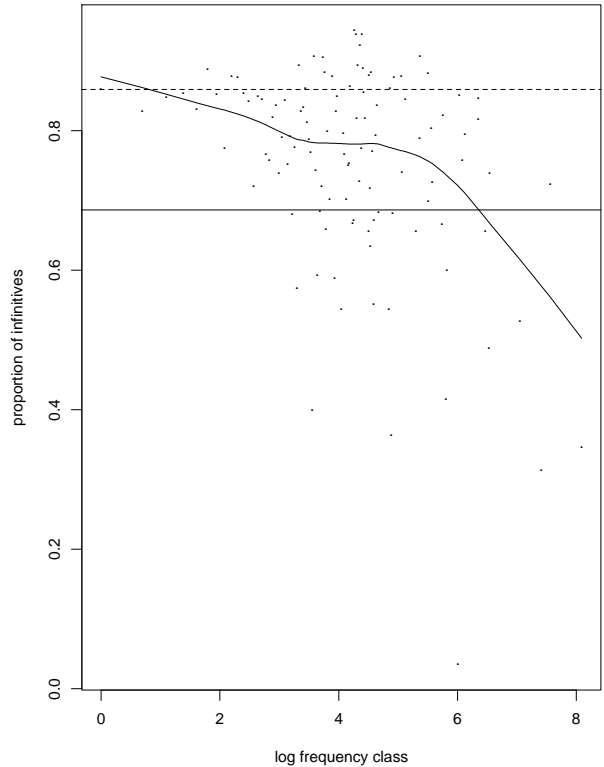

Figure 1: Relative frequency of Dutch infinitives versus finite plural in the Uit den Boogaart corpus, as a function of log-frequency class. The horizontal solid line represents the overall MLE, the relative frequency of the infinitive as computed over all tokens; the horizontal dashed line represents the relative frequency of the infinitive among the hapax legomena. The solid curve represents a non-linear smoothing using running medians.

Figure 1 shows that this prediction is not borne out. This scatterplot shows the relative frequency of the infinitive versus the finite plural, as a function of the log-frequency of the *-en* form. At the left-hand edge of the graph, the relative frequency of the infinitives for the hapax legomena is shown. This proportion is also highlighted by the dashed horizontal line. As we proceed to the right, we observe that there is a general downward curvature representing a lowering of the proportion of the infinitives for the higher frequency words. This trend is captured by the solid non-parametric regression line. The solid *horizontal* line represents the proportion of infinitives calculated over *all* frequency classes. The two horizontal lines can be interpreted as MLEs for the probability of an *-en* form being an infinitive: the solid line or *overall* MLE clearly provides an estimate based on the whole population, whereas the dashed line or *hapax-based* MLE provides an estimate for the hapaxes. The overall MLE computes a lower relative frequency for the infinitives, compared to the hapax-based MLE. The question then is which of these MLEs provides a better estimate for low frequency types. In particular, for types that have not been seen in the training corpus, and for which we therefore have no direct estimate of the word-specific prior probabilities, we would like to know whether the hapax-based or overall MLE provides a better estimate. In the light of the fact that the probability of observing a novel instance of a morphological construction is well estimated by the proportion of hapaxes instantiating that construction to the total number of hapaxes (see (Baayen, 1993)), we have some reason to expect the hapax-based MLE to be more accurate.

To test this expectation we compared the accuracy of the overall and hapax-based MLEs using 10-fold cross-validation. We first randomized the list of *-en* tokens from the UdB corpus, and divided this into ten equal-sized parts. Each of the ten parts was held out as the test set, and the remaining nine tenths was used as the training set over which the two MLE estimates were computed. The results are shown in Table 1. In this table, $N_0(inf)$ and $N_0(pl)$ represent the observed number of tokens of infinitives and plurals in the held-out portion of the data, representing types that had not been seen in the training data. The final four rows compare the estimates for these numbers of tokens given the overall MLE ($E_o(N_0(inf))$ and $E_o(N_0(pl))$), versus the hapax-based MLE ($E_h(N_0(inf))$ and $E_h(N_0(pl))$). For all ten runs, the hapax-based MLE is clearly a far better predictor than the overall MLE.[1]

The pattern that we have observed for the Dutch infinitive-plural ambiguity can be replicated for other cases of morphological ambiguity. Consider the case of English verbs ending in *-ed*, which are systematically ambiguous between being simple past tenses and past participles. The upper panel of Figure 2 shows the distribution of the relative frequencies of the two functions, plotted against log frequency class for the Brown corpus (Francis and Kucera, 1982). (All lines, including the non-parametric regression line are interpretable as in Figure 1.) Results of a ten-fold cross validation are shown in Table 2. Although the magnitude of difference between the overall MLE and the hapax-based MLE is smaller than in the previous example, nonetheless the hapax-based MLE is still a significantly better predictor.[2]

---

[1] A paired t-test on the ratios $N_0(inf)/N_0(pl)$ versus $E_o(N_0(inf))/E_o(N_0(pl))$ reveals a highly significant difference ($t_9 = 13.4, p < 0.001$); conversely a comparison of $N_0(inf)/N_0(pl)$ and $E_h(N_0(inf))/E_h(N_0(pl))$ reveals no difference ($t_9 = 0.96, p > 0.10$).

Table 1: Results of 10-fold cross-validation for Dutch *-en* verb forms from the Uit den Boogaart corpus. Columns represent different cross-validation runs. *N(inf)* and *N(pl)* are the number of tokens of the infinitives and finite plurals, respectively, in the training set. $N_1(inf)$ and $N_1(pl)$ are the the number of tokens of the infinitives and finite plurals, respectively, among the hapaxes in the training set. *OMLE* and *HMLE* are, respectively, the overall and hapax-based MLEs. $N_0(inf)$ and $N_0(pl)$ denote the number of tokens in the held-out portion that have *not* been observed in the training set. The expected numbers of tokens of infinitives and plurals for types unseen in the training set, using the overall MLE are denoted as $E_o(N_0(inf))$ and $E_o(N_0(pl))$; the corresponding estimates using the hapax-based MLE are denoted as $E_h(N_0(inf))$ and $E_h(N_0(pl))$.

| Run | 1 | 2 | 3 | 4 | 5 | 6 | 7 | 8 | 9 | 10 |
|---|---|---|---|---|---|---|---|---|---|---|
| N(inf) | 19509 | 19527 | 19536 | 19526 | 19507 | 19511 | 19533 | 19524 | 19569 | 19585 |
| N(pl) | 8953 | 8935 | 8926 | 8936 | 8955 | 8952 | 8930 | 8939 | 8894 | 8878 |
| OMLE | 0.685 | 0.686 | 0.686 | 0.686 | 0.685 | 0.685 | 0.686 | 0.686 | 0.688 | 0.688 |
| $N_1$(inf) | 1075 | 1086 | 1066 | 1068 | 1092 | 1091 | 1098 | 1066 | 1094 | 1079 |
| $N_1$(pl) | 185 | 184 | 180 | 182 | 179 | 185 | 184 | 178 | 179 | 180 |
| HMLE | 0.853 | 0.855 | 0.856 | 0.854 | 0.859 | 0.855 | 0.856 | 0.857 | 0.859 | 0.857 |
| $N_0$(inf) | 120 | 114 | 133 | 125 | 133 | 123 | 102 | 118 | 121 | 127 |
| $N_0$(pl) | 24 | 19 | 20 | 18 | 18 | 16 | 15 | 23 | 23 | 21 |
| $E_o(N_0$(inf)) | 99 | 91 | 105 | 98 | 103 | 95 | 80 | 97 | 99 | 102 |
| $E_o(N_0$(pl)) | 45 | 42 | 48 | 45 | 48 | 44 | 37 | 44 | 45 | 46 |
| $E_h(N_0$(inf)) | 123 | 114 | 131 | 122 | 130 | 119 | 100 | 121 | 124 | 127 |
| $E_h(N_0$(pl)) | 21 | 19 | 22 | 21 | 21 | 20 | 17 | 20 | 20 | 21 |

In the two examples we have just considered, the hapax-based MLE, while being a better predictor of the a priori lexical probability for unseen cases than the overall MLE, does not actually yield a different prediction as to which function of a form is more likely. This does not hold generally, however, and the bottom panel of Figure 2 presents a case where the hapax-based MLE actually yields a different prediction as to which function is more likely. In this plot we consider Dutch word forms from the UdB corpus ending in *-en*. Dutch *-en* is used as a marker in verbs, as we have seen, where it marks the infinitive, present plural, and also the past plural for strong verbs; it is also used as a marker of noun plurals. This case is somewhat different from the preceding two cases since it is not strictly speaking a case of morphological syncretism. However, it is a potential source of ambiguity in text analysis, since a low frequency form in *-en*, where one may not have seen the stem of the word, could potentially be either a noun or a verb. Also, systematic ambiguity exists among cases of noun-verb conversion: for example *fluiten* is either a noun meaning 'flutes' or a verb meaning 'to play the flute'; *spelden* means either 'pins' or 'to pin'; and *ploegen* means either 'ploughs' or 'to plough'. Results for a ten-fold cross validation for these data are shown in Table 3.[3] In this case, the overall MLE would lead one to predict that for an unseen form in *-en*, the verbal function would be more likely. Contrariwise, the hapax-based MLE predicts that the nominal function would be more likely. Again, it is the hapax-based MLE that proves to be superior.

---

[2] A paired t-test on the ratios $N_0(vbn)/N_0(vbd)$ versus $E_o(N_0(vbn))/E_o(N_0(vbd))$ reveals a significant difference ($t_9 = 2.47, p < 0.05$); conversely a comparison of $N_0(vbn)/N_0(vbd)$ and $E_h(N_0(vbn))/E_h(N_0(vbd))$ reveals no difference ($t_9 = 0.48, p > 0.10$).

Table 2: Cross validation statistics for English past participles versus simple past tense verbs.

| Run | 1 | 2 | 3 | 4 | 5 | 6 | 7 | 8 | 9 | 10 |
|---|---|---|---|---|---|---|---|---|---|---|
| N(inf) | 20386 | 20360 | 20376 | 20372 | 20388 | 20451 | 20431 | 20431 | 20426 | 20400 |
| N(pl) | 13845 | 13871 | 13855 | 13859 | 13843 | 13781 | 13801 | 13801 | 13806 | 13832 |
| OMLE | 0.596 | 0.595 | 0.595 | 0.595 | 0.596 | 0.597 | 0.597 | 0.597 | 0.597 | 0.596 |
| $N_1$(vbn) | 701 | 695 | 678 | 700 | 693 | 705 | 690 | 692 | 710 | 711 |
| $N_1$(vbd) | 395 | 401 | 405 | 406 | 406 | 403 | 404 | 405 | 393 | 403 |
| HMLE | 0.640 | 0.634 | 0.626 | 0.633 | 0.631 | 0.636 | 0.631 | 0.631 | 0.644 | 0.638 |
| $N_0$(vbn) | 80 | 86 | 101 | 83 | 71 | 61 | 85 | 75 | 72 | 77 |
| $N_0$(vbd) | 49 | 52 | 37 | 41 | 43 | 45 | 41 | 50 | 48 | 42 |
| $E_o(N_0$(vbn)) | 77 | 82 | 82 | 74 | 68 | 63 | 75 | 75 | 72 | 71 |
| $E_o(N_0$(vbd)) | 52 | 56 | 56 | 50 | 46 | 43 | 51 | 50 | 48 | 48 |
| $E_h(N_0$(vbn)) | 83 | 88 | 86 | 78 | 72 | 67 | 79 | 79 | 77 | 76 |
| $E_h(N_0$(vbd)) | 46 | 50 | 52 | 46 | 42 | 39 | 47 | 46 | 43 | 43 |

---

[3] A paired t-test on the ratios $N_0(v)/N_0(n)$ versus $E_o(N_0(v))/E_o(N_0(n))$ reveals a highly significant difference ($t_9 = 95.95, p < 0.001$); conversely a comparison of $N_0(v)/N_0(n)$ and $E_h(N_0(v))/E_h(N_0(n))$ reveals no difference ($t_9 = 0.12, p > 0.10$).

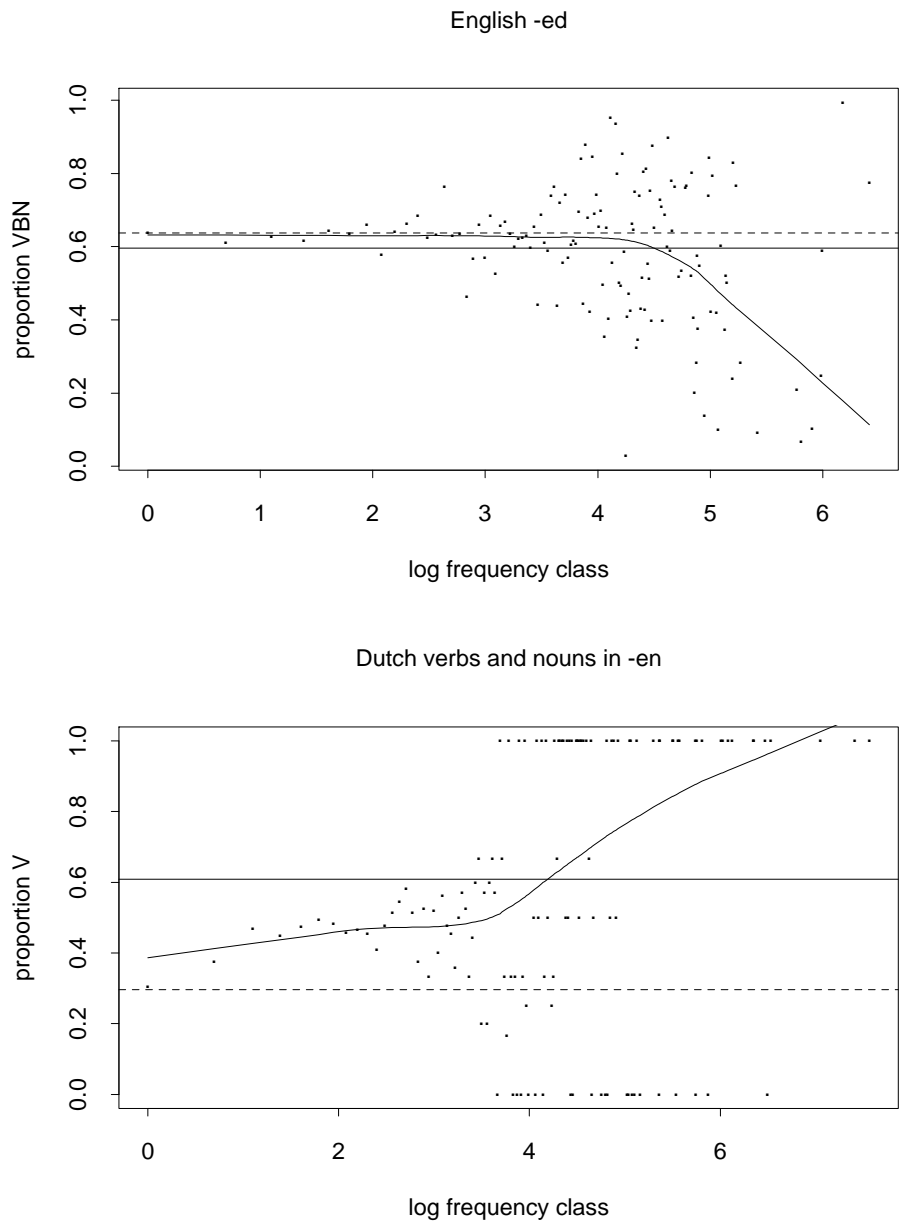

Figure 2: The top panel displays the distribution in the Brown corpus of the relative frequencies of English simple past tense verbs in *-ed* (Brown corpus tag *VBD*) versus past participles in *-ed* (*VBN*), plotted against log frequency class. The bottom panel displays the relative frequency as a function of frequency class of Dutch verbs in *-en* (infinitives, present plurals and strong past tense plurals), versus plural nouns in *-en*, computed over the Uit den Boogaart corpus.

Table 3: Cross validation statistics for Dutch verbs in *-en* versus plural nouns in *-en*.

| Run | 1 | 2 | 3 | 4 | 5 | 6 | 7 | 8 | 9 | 10 |
|---|---|---|---|---|---|---|---|---|---|---|
| N(v) | 25237 | 25283 | 25267 | 25245 | 25292 | 25267 | 25205 | 25207 | 25261 | 25294 |
| N(n) | 18306 | 18260 | 18277 | 18299 | 18252 | 18277 | 18339 | 18337 | 18283 | 18250 |
| OMLE | 0.580 | 0.581 | 0.580 | 0.580 | 0.581 | 0.580 | 0.579 | 0.579 | 0.580 | 0.581 |
| $N_1(v)$ | 1312 | 1295 | 1287 | 1317 | 1284 | 1298 | 1298 | 1297 | 1292 | 1298 |
| $N_1(n)$ | 2913 | 2910 | 2939 | 2942 | 2901 | 2922 | 2979 | 2969 | 2936 | 2931 |
| HMLE | 0.311 | 0.308 | 0.305 | 0.309 | 0.307 | 0.308 | 0.303 | 0.304 | 0.306 | 0.307 |
| $N_0(v)$ | 124 | 131 | 154 | 142 | 148 | 143 | 148 | 156 | 153 | 139 |
| $N_0(n)$ | 325 | 344 | 327 | 334 | 352 | 335 | 289 | 301 | 327 | 319 |
| $E_o(N_0(v))$ | 260 | 276 | 279 | 276 | 290 | 277 | 253 | 265 | 278 | 266 |
| $E_o(N_0(n))$ | 189 | 199 | 202 | 200 | 210 | 201 | 184 | 192 | 202 | 192 |
| $E_h(N_0(v))$ | 139 | 146 | 146 | 147 | 153 | 147 | 133 | 139 | 147 | 141 |
| $E_h(N_0(n))$ | 310 | 329 | 335 | 329 | 347 | 331 | 304 | 318 | 333 | 317 |

## 3 Discussion

As we have seen in the three examples discussed above, the MLE computed over hapax legomena yields a better prediction of lexical prior probabilities for unseen cases, than does an MLE computed over the entire training corpus. We now have to consider why this result holds. As we shall see, the reasons are different from case to case, but nonetheless share a commonality.

Consider first the last set of data, namely the ambiguity in Dutch between *-en* verb forms and *-en* plural nouns. Ceteris paribus, plural nouns are less frequent than singular nouns; on the other hand, *-en* for verbs serves both the function of marking plurality and of marking the infinitive. High frequency verbs include some very common word forms, such as auxiliaries like *hebben* 'have', *zullen* 'will', *kunnen* 'can' and *moeten* 'must'. Thus, for the high frequency ranges, the data is weighted heavily towards verbs. On the other hand, while both nouns and verbs are open classes, nouns are far more productive as a class than are verbs (Baayen and Lieber, 1991), and this pattern becomes predominant in the low frequency ranges: among low frequency types, most tokens are nouns. Hence, for the low frequency ranges, the data is weighted towards nouns. These two opposing forces conspire to yield a downward trend in the percentage of verbs as we proceed from the high to the low frequency ranges.

Next, consider the English past tense versus past participle ambiguity. One of the important functions of the past participle is as an adjectival modifier or predicate; for example, *the parked car*. In this function the past participle has a passive meaning. For reasons that are not entirely clear to us, a predominant number of the high frequency verbs are relatively infelicitous in this usage; these verbs include cases like *walk* and *appear* which are surface intransitives, for which one would not expect to find the passive usage, but they also include transitives like *move*, *try*, *ask* which are not generally felicitous in this usage. Among the low-frequency verbs, including *accentuate*, *bottle* and *incense*, types where the past participle is preferred predominate. What is clear from the plot in the top panel of Figure 2, is that the downward trend in the regression curve to the right of the plot, is due to the lexical properties of a relatively small number of high frequency verbs. For the greater part of the frequency range, there is a relatively stable proportion of participles to finite past forms. Thus, the hapax-based MLE yields an estimate that is uncontaminated by the lexical properties of individual high frequency forms.

Finally, consider the case that we started with of Dutch verb forms in *-en*. In Figure 1 the strong downward trend in the regression curve at the right of the figure is due in large measure to the inclusion of high frequency auxiliary verbs, examples of which have already been given. These verbs, while possible in the infinitival form, occur predominantly in the finite forms. Hence, a form such as *hebben* 'have' is much more likely to be a plural finite form than it is to be an infinitive. At the low end of the frequency spectrum, we find a great many verbs derived with separable particles, such as *afzeggen* 'cancel'; note that separable prefixation is the most productive verb forming process in Dutch. In the infinitival form, the particle is always attached to the verb. However, in the finite forms in main clauses, the particle must be separated: e.g., *wij* **zeggen** *onze afspraak* **af** 'we are cancelling our appointment'. These properties of Dutch separable verbs boost the likelihood of infinitival forms for the low frequency ranges, but they also boost the likelihood of finite plural forms for the higher frequency forms. Note that the separated form *zeggen* is identical to the underived verb *zeggen* meaning 'say', hence any separated finite forms will accrue to the frequency of the generally much more common derivational base.

What all three of these cases share, is that the statistical properties of the high-frequency ranges are dominated by lexical properties of particular sets of high-frequency words. This in turn biases the overall MLE, and makes it a poorer predictor of novel cases, than the hapax-based MLE, which is less sensitive to the idiosyncratic lexical properties to which high-frequency words are often subject. Of course, low frequency classes may also display relevant lexical biases, as was the case with the Dutch separable particle verbs that we have just discussed. However, the lexical properties displayed in such cases are more likely to be representative of novel cases than the lexical properties displayed for the high frequency forms. This is because if a lexical property is important enough to dominate the low frequency classes, it must hold over a large number of types, and hence is very likely to represent a productive process in the language. This ties in directly with the observation that highly productive morphological processes give rise to large numbers of very low frequency types (Baayen, 1989).

The results of the analyses presented in this paper have important consequences for the development of tagging systems. For high frequency words, one can obtain fairly reliable estimates of the lexical priors by tagging a corpus that gives a good coverage to words of various ranges. For predicting the lexical priors for the much larger mass of very low frequency types, most of which would not occur in any such corpus, the results we have presented suggest that one should concentrate on tagging a good representative sample of the hapaxes, rather than extensively tagging words of all frequency ranges.

Note that the techniques introduced here are useful not only in computing lexical priors for forms that are in all respects formally identical, but also for forms which may be identical only, for instance, in their written form. For example, the written form *goroda* in Russian (Cyrillic города) may either be the nominative plural or the genitive singular of 'city'. In the genitive singular, the stress is on the first syllable (/gˈorədə/), whereas in the nominative plural the stress resides on the final syllable (/garadˈa/). This stress difference is important in such applications as text-to-speech, since not only does stress placement affect the position of pitch accents, but it also has a radical affect on vowel quality in Russian.